\documentclass[sigconf]{acmart}

\copyrightyear{2023}
\acmYear{2023}
\setcopyright{acmlicensed}\acmConference[GECCO '23 Companion]{Genetic
and Evolutionary Computation Conference Companion}{July 15--19,
2023}{Lisbon, Portugal}
\acmBooktitle{Genetic and Evolutionary Computation Conference Companion
(GECCO '23 Companion), July 15--19, 2023, Lisbon, Portugal}
\acmPrice{15.00}
\acmDOI{10.1145/3583133.3596410}
\acmISBN{979-8-4007-0120-7/23/07}

\usepackage{adjustbox,afterpage,amsfonts,array,booktabs,dsfont}
\usepackage{float,graphicx,hyperref,lscape,multibib,multirow}
\usepackage{pgfplots,pifont,rotate,soul,subfigure,xcolor,balance}
\pgfplotsset{compat=1.17}

\begin{document}
%\title[Analysis of BCHM in Conjunction with DE Performance on SBOX-COST Suite]{Analysis of Bound Constraints Handling Methods in Conjunction with Differential Evolution Performance on SBOX-COST Benchmarking Suite}
\title[Patterns of convergence and bound constraint violation in DE on SBOX-COST]{Patterns of Convergence and Bound Constraint Violation in Differential Evolution on SBOX-COST Benchmarking Suite%: towards adaptive bound-constraints handling
}

\author{Mădălina-Andreea Mitran}
\affiliation{%
      \institution{West University of Timisoara}
      \streetaddress{blvd. Vasile Pârvan, 4}
      \city{Timisoara}
      \postcode{300223}
      \country{Romania}}
\email{madalina.mitran96@e-uvt.ro}
\orcid{0000-0002-7946-7054}

\author{Anna V. Kononova}
\affiliation{%
      \institution{LIACS, Leiden University}
      \streetaddress{Niels Bohrweg 1}
      \city{Leiden}
      \postcode{2333}
      \country{The Netherlands}}
\email{a.kononova@liacs.leidenuniv.nl}
\orcid{0000-0002-4138-7024}

\author{Fabio Caraffini}
\affiliation{%
      \institution{Swansea University}
      \streetaddress{Fabian Way}
      \city{Swansea}
      \postcode{SA1 8EN}
      \country{UK}}
\email{fabio.caraffini@swansea.ac.uk}
\orcid{0000-0001-9199-7368} 

\author{Daniela Zaharie}
\affiliation{%
      \institution{West University of Timisoara}
      \streetaddress{blvd. Vasile Pârvan, 4}
      \city{Timisoara}
      \postcode{300223}
      \country{Romania}}
\email{daniela.zaharie@e-uvt.ro}
\orcid{0000-0003-3388-6058}

\renewcommand{\shortauthors}{Mitran et al.}

\begin{abstract}
% \fc{
This study investigates the influence of several bound constraint handling methods (BCHMs) on the search process specific to Differential Evolution (DE), with a focus on identifying similarities between BCHMs and grouping patterns with respect to the number of cases when a BCHM is activated. The empirical analysis is conducted on the SBOX-COST benchmarking test suite, where bound constraints are enforced on the problem domain. This analysis provides some insights that might be useful in designing adaptive strategies for handling such constraints. 
%}
\end{abstract}




\maketitle

\section{Introduction}
% motivation - current awareness of the role of BCHM
%\dz{
For most real-world optimisation problems, the optimum should be searched in a region limited by a boundary because outside this region the objective function might not be evaluable. Ensuring that the search is conducted within the bounds requires strategies for dealing with the candidate solutions which are outside the bounds, strategies which are commonly called bound constraint handling methods (BCHM)\footnote{BCHM is also referred to as a strategy of dealing with infeasible solutions (SDIS)~\cite{Kononova2022importance}.}. The influence of a BCHM on the performance of different population-based metaheuristics has been studied in several works (Covariance Matrix Adaptation Evolutionary Stra\-tegy \cite{Wessing2013, Biedrzycki2020}, Particle Swarm Optimisation \cite{Helwig2013, Castillo2017, Oldewage2018}, Differential Evolution \cite{Arabas2010, Padhye2015, Biedrzycki2019, Boks2021, KONONOVA2021, Kadavy2022}). The importance of considering BCHM as a specific algorithmic component that could interfere with the behaviour of the metaheuristic algorithm has recently been acknowledged in \cite{Kadavy2022, Kononova2022importance} where the specification of the used BCHM is considered critical for the unambiguous reproducibility of the results.%} 

%\dz{
Despite the rather large number of BCHMs analysed in the papers mentioned above, the current knowledge on the role of a BCHM is rather limited, and it is not yet clear to which extent a BCHM has also a guidance role in the search process, beyond just bringing back the candidate solutions to the feasible region.  The influence of a BCHM on the optimisation process depends, at least, on how frequently it is applied and on how its way of action interferes with the variation operators of the search algorithm. The frequency of applying a BCHM depends on the number of infeasible candidate solutions, which is related, in its turn, to several factors: fitness landscape, location of the optima, problem di\-men\-sio\-na\-lity and size of the search neighbourhood defined by the variation operators. On the other hand, some BCHM mechanisms preserve the current population distribution and/or the search direction induced by the variation operators, while other BCHM mechanisms generate elements that, otherwise, could not be easily generated by the variation operators~\cite{Kononova2022importance}.%} 

% about benchmarks for bound-constrained problems
%\dz{
Due to all these entangled factors, it is not easy to decipher the reasons why a BCHM is effective or not in conjunction with some search algorithms for some optimisation problems. With some exceptions \cite{Helwig2008, Ali2006}, most studies are empirical, based on public benchmarking suites such as CEC~\cite{CEC} and BBOB \cite{BBOB2021}. Most analyses use CEC test functions, e.g. CEC2006 \cite{Juarez-Castillo2019}, CEC2017~\cite{Biedrzycki2019, Kadavy2022, kreischer2017}, CEC2020~\cite{Kadavy2022} and only a couple of them use BBOB \cite{Boks2021, Kononova2022importance}. This might be explained by the fact that in the original BBOB test suite, for most of the function instances, the optimum is not near the boundary; thus the BCHM is activated only in the first stages of the search, thus having only a transient effect. The newly proposed SBOX-COST variant of BBOB\footnote{accessible through \texttt{IOHProfiler}~\cite{IOHprofiler}} is characterised by the fact that it allows the generation of function instances with the global optimum arbitrarily close to the boundary, unlike the original BBOB where all functions except linear slope ($F5$) have a wide outside perimeter free of optima on all possible instances. However, it should be mentioned that such modification has not been implemented for functions $F4$, $F5$, $F8$, $F9$, $F19$, $F20$, $F24$ in SBOX-COST - their optima locations are kept in $[-4,4]^n$ (as in BBOB). Domain of all functions in SBOX-COST is identical to BBOB, $[-5,5]^n$.
%}
%\ak{does it also matter that it evaluates points outside domain to $\infty$?}

%\ak{mention BBOB's instance generating functionality that allows varying locations of optima within the inner subdomain}
%\dz{
The aim of this paper is to investigate the influence of se\-ve\-ral bound constraints handling methods on the search process specific to Differential Evolution. %, with a focus on adaptive variants.  
The empirical analysis is conducted on the SBOX-COST suite and addresses the following questions:
\begin{itemize}
\item Which SBOX-COST functions are characterised by a landscape that favours the search near the boundary, i.e. ge\-ne\-ra\-tion of infeasible solution candidates (Section 4)?
\item Is it beneficial to preserve, by the bound constraint handling method, some DE search characteristics, e.g. population mean and variance, search direction (Section 5)?
\item Is it beneficial to incorporate into the BCHM information from some feasible reference elements, e.g. best element in the population, mean of the population (Section 5)?
\item How can a pool of BCHMs be constructed in the context of adaptive selection of the correction (Section 6)? 
\end{itemize}
%}

%\dz{The rest of the paper is organised as follows ...}
%\ak{please cite \cite{Kononova2022importance} somewhere} - %\dz{done - please check} 
%\ak{don't you want to refer more to our results in that paper?}
%\dz{\hl{@Anna}: please include references to related results}
%\ak{we need to agree on what terminology we use: SDIS or BCHM. I don't mind either, maybe we could just mention they are equivalent?} \dz{I would suggest BCHM - is more popular)} \ak{ok!}

\section{Framework of the analysis}
% benchmark
% questions
% -  it is useful to preserve the search characteristics (e.g. population mean and variance - beta, search direction - vector target)
% - interference between correction and parameter adaptation 
%\dz{
The analysis is carried out in the context of minimising a function $f:D\subset \mathbf{R}^n\to \mathbf{R}$, with $D=[a_1,b_1]\times [a_2,b_2]\times \ldots \times [a_n,b_n]$, $a_i<\infty$, $b_i<\infty$, under the assumption that all candidate solutions should belong to $D$. As the behaviour of a BCHM is correlated with the search metaheuristic, we will focus on only one metaheuristic, e.g. Differential Evolution.
%} 

\subsection{Differential Evolution variants}

%\fc{
There is an established jargon when it comes to describing the DE framework and its variants \cite{bib:DEbook}. As in an evolutionary algorithm, in DE, a set of $N$ $n$-dimensional candidate solutions $x^j$ ($j=1,2,\dots, N$), called individuals, form a population. In the generational loop, for each $j^{th}$ individual, called \texttt{target}, a \texttt{trial} individual, $\mathbf{y}^j$, is constructed by mutation and crossover. Once all \texttt{trial} individuals are constructed, each of them will replace the corresponding \texttt{target} individual if it has a smaller value of the objective function than that of the \texttt{target} individual. %In the generational loop, each $j^{th}$ individual is selected to undergo variation operators, thus producing an offspring solution $\mathbf{y}^j$ called \texttt{trial}. This \texttt{trial} can deterministically replace the position of the perturbed $j^{th}$ solution originating it, called \texttt{target} in the DE context, based on the value of its objective function. %However, in the original DE framework, replacement is not a greedy process and occurs only after one entire generation, that is, when all (\texttt{target}) solutions have been perturbed from the first to the last to obtain $N$ \texttt{trial} solutions. 
%Generations are repeated following the same selection order $j=1,2,\dots, N$ of the \texttt{target} points. For each \texttt{target}, an auxiliary \texttt{mutant} solution $\mathbf{x_m}$ is first generated by 
Mutation consists in linearly combining existing individuals. % with a `mutation' operator. 
Many implementations exist for this operator; see \cite{DAS20161} for details. All established mutation operators require a scale factor $F\in[0,2]$ and are characterised by the notation \texttt{A/B}. Here, \texttt{A} can be the \texttt{target} individual (or \texttt{current}), the \texttt{best} so far solution, a randomly chosen individual from the population (i.e., \texttt{rand}), or a computed vector of the kind \texttt{current-to-best}, etc. 
%In the latter case, since \texttt{target} has already been (arithmetically) recombined with another individual within the mutation operator, it is common (but not necessary) to skip a further recombination step that must be performed otherwise. 
To move \texttt{A} within the search space, a number \texttt{B} of `difference vectors' (usually $B=1$ or $B=2$) is computed as the ($F$-) scaled difference between different randomly chosen individuals and added to \texttt{A}. Subsequently, a crossover operator \texttt{C} is fed with the \texttt{target} and \texttt{mutant} individuals to generate a \texttt{trial}. Note that the \texttt{mutant} might contain infeasible components (the mutation in DE is the only operator that can generate infeasible solutions). As some of these can be transferred to \texttt{trial} during the crossover/recombination process, a BCHM is needed. Therefore, a complete DE variant can be indicated with the notation \texttt{DE/A/B/C-BCHM}. The most widely used crossover operators for DE are referred to as \texttt{bin} and \texttt{exp} strategies, which stand for binomial and exponential, respectively. Note that a prefixed value for the `crossover rate' parameter $CR\in[0,1)$ is needed in both crossover strategies. However, $CR$ assumes a significantly different meaning when \texttt{C=bin}, where it determines the probability of exchanging each component %(or, in other words, the average number of components to be exchanged), 
and when \texttt{C=exp}, where $n$ and $CR$ decide the length of a burst of consecutive components to be exchanged \cite{ISBDE}. In the latter case, the probability of exchanging the next component follows a geometrical progression, thus decaying exponentially. We refer to \cite{bib:DEbook,DAS20161} for more details on crossover strategies.%} 

%\fc{
The \texttt{DE/rand/1/bin-$\star$} (with $\star$ indicating a generic BCHM) configuration represents the original DE framework. Advances in DE have produced modern variants that feature new variation operators, additional operators, hybrid algorithmic components, and, most importantly, various self-adaptation mechanisms for the control parameters $F$ and $CR$, as well as for adjusting the population size $N$. %} 
%\fc{
Achieving the optimal self-adaptive algorithm is still an open question in heuristic optimisation, and it is key in DE, given its sensitivity to the parameter setting (despite having a reasonably small number of parameters with respect to other heuristics). 
%}
%
%
% \subsubsection{State-of-the-art}
% % \fc{
% % SHADE\\
% % LSHADE\\
% % Shall I describe them here? or briefly withtou subsubsection?}
% % DE/rand/1/bin, SHADE, LSHADE
% \fc{Almost 3 decades of research in DE have produced numerous DE variants. In most cases, these still follow to some degree the original framework and embed a self-adaptive logic in the attempt of achieving self-adaptation of $F$, $CR$ and N to the specific problem at hand.}

%\fc{
The current state-of-the-art is represented by self-adaptive DE frameworks such as the Success-History based Adaptive DE (\texttt{SHADE}) \citep{bib:tanabe2013}, and its successor \texttt{L-SHADE} \cite{tanabe2014improving}, which builds on the success of \texttt{SHADE} by adding a linear rule to reduce the size of the population during the optimisation process. In this light, while the first algorithm only self-tunes $F$ and $CR$, the second algorithm dynamically changes $N$ during the optimisation process from an initial suggested value of $N=18\cdot n$. Apart from this difference, these two advanced DE algorithms use the same `\texttt{current-to-\textit{p}best/1}' mutation strategy proposed in \citep{bib:zhang2009} for the popular \texttt{JADE} algorithm (where the \textit{p}best vector is selected from the $p\%$ best individuals in the population), as well as a memory-based system to adapt the control parameters. The latter consists of storing both the weighted Lehmer average of successful $F$ values and the weighted arithmetic average of successful $CR$ values from previous generations. When needed, such values are randomly selected from the memory. Note that the value of \textit{p} is not prefixed as in \texttt{JADE}, but is generated randomly. This mutation operator is usually followed by the binomial crossover. Therefore, the structure of these algorithms can be seen as a  \texttt{DE/current-to-\textit{p}best\-/1/bin} algorithm with a memory-based adaptation system for control parameters and, in the case of \texttt{L-SHADE}, a population size reduction mechanism (which is shown to be beneficial in previous DE variants such as \cite{bib:Brest2008b}). Despite these simple ideas, \texttt{SHADE} and \texttt{L-SHADE} have shown great performances in several benchmark problems compared to other DE algorithms.%} %$18\cdot n$

\begin{table*}
  \caption{Bound constraint handling methods ($c_L$ denotes correction for components violating the lower bound; $c_U$ denotes correction for components violating the upper bound; $c$ denotes a correction which does not depend on the violated bound). %\ak{does $c(y_i)$ refer to both $c_L$ and $c_U$?}
  }
  \label{tab:BCHM}
  \begin{tabular}{ccccc}
    \toprule
    \multirow{2}{*}{Method}& Corrected & Application & \multirow{2}{*}{Type} & \multirow{2}{*}{Characteristics}\\
    & component/element & level & &\\
    \midrule
    \multirow{2}{*}{\texttt{Saturation}} & $c_L(y_i)=a_i$ & \multirow{2}{*}{component} & \multirow{2}{*}{deterministic} & \multirow{2}{*}{set on the violated bound}\\
    & $c_U(y_i)=b_i$ & & & \\
    \multirow{2}{*}{\texttt{Mirror}} & $c_L(y_i)=2a_i-y_i$ & \multirow{2}{*}{component} & \multirow{2}{*}{deterministic} & \multirow{2}{*}{a larger violation leads to a larger correction}\\
    & $c_U(y_i)=2b_i-y_i$ & & & \\
    \texttt{Uniform} &   $c(y_i)\sim\mathcal{U}(a_i,b_i)$& component     &  stochastic & $c(y_i)$ is independent of the position of $y_i$  \\
    \texttt{Beta}  & $c(y_i)\sim a_i+Beta(\alpha_i,\beta_i)(b_i-a_i)$& component     &  stochastic &   aims to preserve population mean and variance\\
    \texttt{ExpC}  & Eq.(\ref{eq:expC}) & component     &  stochastic &  it uses a reference point \\
    \multirow{2}{*}{\texttt{Vector}} & $c(y)=\alpha y+(1-\alpha)R$& \multirow{2}{*}{vector} & \multirow{2}{*}{deterministic} & it uses a reference point  \\
      & $\alpha\in (0,1)$ as in Eq.(\ref{eq:vectorAlpha}) &      &   &  (when $R=x$ it preserves the search direction) \\
  \bottomrule
\end{tabular}
\end{table*}

\subsection{Bound constraint handling methods}
%\dz{
An infeasible trial individual, $y$, contains at least one out-of-bounds component (i.e. $\exists i\in \{1,\ldots,n\} : y_i\not\in [a_i,b_i]$). A bound constraint handling method transforms all out-of-bounds components (in the case of {\it component-wise} methods) or all components, including the feasible ones (in the case of {\it vector-wise} methods) in such a way that the resulting individual becomes a feasible one.

There are various ways of transforming an infeasible indivi\-dual into a feasible one. These are different with respect to what information about the infeasible individual they use (e.g. which bound is violated, how large is the violation of the bound) and with respect to the nature of the strategy of generating values inside the bounding box (e.g. deterministic or stochastic). There are more than fifteen BCHMs that have been included in various empirical studies. Unfortunately, the terminology is not standardized, thus the same BCHM is referred to by using different names (e.g. the BCHM replacing the infeasible component with the closest bound is mentioned as \texttt{projection}~\cite{kreischer2017, Juarez-Castillo2019, Biedrzycki2019}, \texttt{clipping}~\cite{Kadavy2022}, \texttt{saturation}~\cite{KONONOVA2021}).%}

%\dz{
The BCHMs included in the current analysis are summarized in Table \ref{tab:BCHM}, where besides some of the popular strategies (\texttt{saturation/} \texttt{projection/} \texttt{clipping}, \texttt{mirror/} \texttt{reflection}, \texttt{uniform/} \texttt{reini\-ti\-alization}) a stochastic me\-thod is included which is designed based on the particularities of DE population distribution (\texttt{Beta}) and variants of some previously proposed methods: exponentially confined \cite{Padhye2015} and a vector-wise `scaled mutant' \cite{kreischer2017}. These new variants are described more in detail in the following.%} 

% beta
\subsubsection{Correction based on Beta distribution}\label{sect:BetaCorrection}
%\dz{
Starting from the result presented in \cite{Ali2006} which states that the distribution of the population of feasible DE trial elements (obtained by using \texttt{DE/rand/1} mutation) can be approximated using a Beta distribution (with values in $[0,1]$), one can design a BCHM that generates feasible elements following this distribution. The main idea behind the Beta correction is to generate feasible components having the same mean and variance as the current population.  To satisfy this condition, the parameters $\alpha_i$ and $\beta_i$ ($i\in \{1,\ldots,n\}$) of the Beta distribution are calculated based on the mean ($\text{Mean}(X_i)$) and the variance ($\text{Var}(X_i)$) of the current population (computed along the infeasible component), as described in Eq.~\ref{eq:betaParam}, where $m_i=(\text{Mean}(X_i)-a_i)/(b_i-a_i)$ and $v_i=\text{Var}(X_i)/(b_i-a_i)^2$. In the case where $m_i=0$ or $m_i=1$ (which could happen if the optimum is at the boundary), then $m_i$ is replaced with $\epsilon$ or $1-\epsilon$, respectively ($\epsilon>0$ is a small value, for example, $\epsilon = 0.1$).
\begin{equation} \label{eq:betaParam}
\alpha_i= m_i\left(\frac{m_i(1-m_i)}{v_i}-1\right), \quad \beta_i=\alpha_i\frac{1-m_i}{m_i}
\end{equation}
%}
\subsubsection{Exponentially confined correction}
% ExpC
%\dz{
This BCHM was proposed in \cite{Padhye2015} as a stochastic correction for which the deviation with respect to the violated bound is computed using an exponential distribution. The calculation details are described in Eq.~\ref{eq:expC} where $y_i$ denotes the infeasible component, $r\sim\mathcal{U}[0,1]$, $r^L_i\sim \mathcal{U}[\exp(a_i-R_i),1]$, $r^U_i\sim \mathcal{U}[\exp(R_i-b_i),1]$, and $R_i$ is the $i^{th}$ component of a feasible reference vector $R$.%}  
\begin{equation}\label{eq:expC}
    c(y_i)=\left\{\begin{array}{ll}
a_i-\ln\left(1+r(e^{a_i-R_i}-1)\right) = a_i-\ln r^L_i & y_i<a_i\\
b_i+\ln \left(1+(1-r)(e^{R_i-b_i}-1)\right)= b_i+\ln r^U_i & y_i>b_i,
\end{array}\right.
\end{equation}
%\dz{
The variant proposed in \cite{Padhye2015} corresponds to the case where $R$ is the \texttt{target} individual. In this paper, we also consider variants when the reference vector is the \textit{p}best element (as defined in the SHADE algorithm) or is the midpoint of the current population (mean of the population elements). These three variants are referred to as  \texttt{ExpTarget}, \texttt{ExpBest}, and \texttt{ExpMidpoint}, respectively.%}  

% Vector
\subsubsection{Vector-wise corrections}
%\dz{
In the context of DE, a correction acting on all components (including feasible ones) as illustrated in Eq.\ref{eq:vector} has been proposed in \cite{kreischer2017} under the assumption that $R=0$ and this null vector belongs to the feasible region. 
\begin{equation} \label{eq:vector}
 c(y)=\alpha y+(1-\alpha)R
\end{equation} 
 In Eq.\ref{eq:vector}, $y$ is the infeasible trial individual (corresponding to the target individual, $x$), $R$ is a reference element from the feasible domain, and $\alpha=\min_{i=1}^n\alpha_{i} \in [0,1]$ with $\alpha_i$ defined in Eq.~\ref{eq:vectorAlpha}.
\begin{equation} \label{eq:vectorAlpha}
    \alpha_{i}=\left\{\begin{array}{ll} (R_i-a_i)/(R_i-y_i) & \hbox{ if } y_i<a_i\\
    (b_i-R_i)/(y_i-R_i) & \hbox{ if } y_i>b_i\\
    1 & \hbox{ if } a_i\leq y_i \leq b_i\\
    \end{array} \right.
\end{equation}%}
%\dz{
In the analysis presented in \cite{Biedrzycki2019} a variant is used in which $R$ is the middle point of the bounding box, i.e. $R=(a+b)/2$.  However, the reference vector $R$ can be anywhere in the feasible domain, as the corrected individual is on the point where the line that joins the infeasible mutant $y$ and the reference point intersects the bounding box $[a_1,b_1]\times \ldots \times [a_n,b_n]$.  Once again, the reference vector can be the target individual (\texttt{VectorTarget}), the best (or \textit{p}best) individual  (\texttt{VectorBest}), the population mean (\texttt{VectorMidpoint}), or an arbitrary individual in the population (not considered in this study). It should be remarked that if $R$ is the target individual, $x$, then the search direction is preserved through correction, since $\cos(y-x,c(y)-x)=\cos(y-x,\alpha (y-x))=1$.%}

\section{Related work}
%\dz{
Several authors analyzed the behaviour of different sets of BCHMs in conjunction with various DE variants leading to various conclusions. Some of these studies are shortly reviewed in the following. %}
%\mm{
Arabas et al \cite{Arabas2010} compared the performance of several strategies, including \texttt{saturation}, \texttt{toroidal}, \texttt{uniform}, and \texttt{mirroring}, and found that the choice of the BCHM could affect the DE performance depending on the characteristics of the problem, such as the position of the optimum and the size of the problem. They observed that \texttt{saturation} and \texttt{mirroring} worked well when the optimum was close to the bounds. They also remarked that for small size problems, the DE performance was not significantly influenced by the used BCHM while for larger size problems, \texttt{saturation} and \texttt{mirroring} were more effective than \texttt{uniform}.%}

%\mm{
Padhye et al \cite{Padhye2015} investigated the effects of two groups of BCHMs, depending on whether the correction is performed vector-wise or component-wise, further divided into deterministic and non-deterministic methods on particle swarm optimisation, genetic algorithms, and differential evolution. They argue that deterministic methods such as \texttt{saturation} tend to reduce population diversity, while non-deterministic methods such as \texttt{uniform} may discard valuable information contained in the current population. On the other hand, when the optimum is located near the center of the feasible domain, no significant differences between different BCHMs were reported, while when it is located close to the boundary of the feasible domain, a vector-wise strategy, referred to as \texttt{inverse-parabolic}, performs the best.%}

%\mm{
The currently most extensive study on the influence of BCHMs on DE performance was presented in \cite{Biedrzycki2019}. Biedrzycki et al analysed the impact of different correction methods on the dynamics of the population, convergence speed, and global optimisation efficiency. They found that the choice of the method could significantly influence the mean and variance of the mutant population distribution, especially for high-dimensional problems. They also observed that adaptive DE variants were less sensitive to the choice of strategy than non-adaptive ones. The best behaviour was induced by strategies \texttt{midpoint} (the corrected component is in the range defined by the corresponding component of the target/base vector and the violated limit) and \texttt{mirroring}. This motivated us to include in the analysis BCHM variants based on a reference vector identical to the target individual. %}

%\mm{
In \cite{kreischer2017}, Kreischer et al recommend \texttt{mirroring} and \texttt{pro\-jec\-tion} to an interior point of the feasible region, followed by \texttt{midpoint target} as strategies which perform well when coupled with DE, in terms of the number of successful runs and the quality of the solutions obtained, in the case of CEC 2017 benchmark.
Martinez et al \cite{Martinez2020} present results on nine different strategies (including \texttt{midpoint target}, \texttt{mirroring}, and \texttt{projection}) applied in the case of four real-world constrained optimisation problems. The main finding is that the effectiveness of the BCHMs depends on the problem, but, in general, \texttt{projection} on the bounds was the most effective.%}

%\mm{
In \cite{Boks2021}, Boks et al. investigated the effect of BCHMs on the performance of DE with a wide range of mutation operators and crossover methods using the BBOB benchmark. They found that no single BCHM was optimal for all configurations and function groups, but recommended \texttt{reinitialization} at least in the case of binomial crossover. \texttt{Resampling} was found to be successful in the majority of the cases considered. This remark motivated us to include in the analysis a method which preserves characteristics of the population distribution, i.e. the proposed \texttt{Beta} correction (see Section \ref{sect:BetaCorrection}). %\texttt{Midpoint-target} in bin configurations was consistently reliable, while \texttt{projection-to-midpoint} was a good second option for exp crossover configurations. 
%}

%\dz{
Addressing the problem that none of the investigated BCHMs is able to deal with particularities of various landscapes and search strategies, in \cite{Juarez-Castillo2019}, Castillo et al proposed an interesting adaptive scheme for selecting the BCHM that is appropriate for a specific search landscape and search stage.%}

%\dz{
Unlike most of the previous works, which focus on the competition between the analysed methods, the current paper aims to also identify similarities between BCHMs as well as grouping patterns with respect to search behaviour.%}

\section{Bound constraint violation patterns}
% Factors that favor the generation of infeasible elements 
% \begin{itemize}
%     \item Search near the boundary (caused by the location near the boundary of low-cost configurations)
%     \item large difference terms involved in the mutation (caused by a dispersed population and/or large values of the scale factor).
% \end{itemize}
%\dz{
The aim of this section is to present an analysis of the similarities between functions belonging to the benchmarking test suites SBOX-COST and BBOB with respect to the number of cases where a BCHM is activated.%}
\subsection{Quantification of bound constraint violation}
%\dz{
The number of cases where a BCHM is activated and interferes with the search process is related to:
(i) the ratio of infeasible trial individuals, i.e. the number of individuals obtained by DE mutation and crossover that contain at least one component which is outside the bounding box, divided by the total number of trial individuals;
(ii) the  ratio of mutated components (included in trial individuals) that are generated out of the bounding box; this ratio is considered in the following as an approximation of the bound violation probability.
In the case of component-wise methods, the amount of BCHM activation depends on the bound violation probability, while in the case of vector-wise methods, it is related to the ratio of infeasible elements.%}

\begin{figure*}[!t]
    \centering
    \includegraphics[width=0.38\linewidth,trim=0mm 6mm 0mm 7mm,clip]{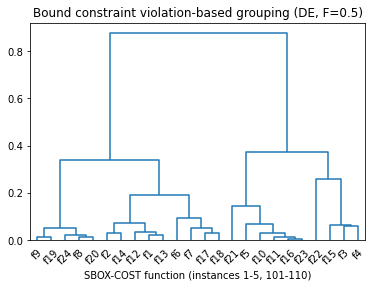}
    \includegraphics[width=0.38\linewidth,trim=0mm 6mm 0mm 7mm,clip]{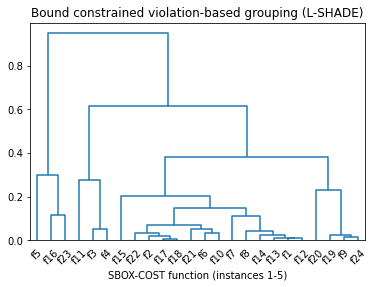}
    \includegraphics[width=0.38\linewidth,trim=0mm 6mm 0mm 7mm,clip]{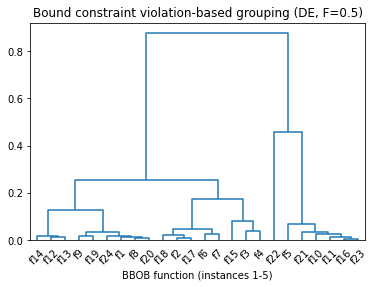}
    \includegraphics[width=0.38\linewidth,trim=0mm 6mm 0mm 7mm,clip]{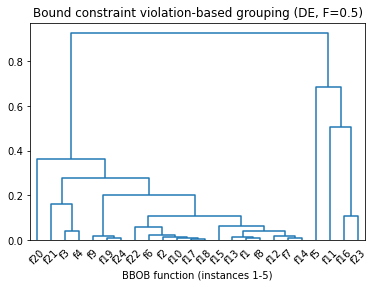}
    \caption{Dendrograms corresponding to agglomerative clustering with complete linkage on cosine similarities between the bound violation probabilities corresponding to all test functions of the suite. Methods: \texttt{DE/rand/1/bin} (left column) and L-SHADE (right column) on SBOX-COST (top row) and BBOB (bottom row) benchmark suites. Instances: 1-5. \label{fig:dendrogram}}
    \Description{Dendrograms illustrating the clustering of SBOX-COST functions}
\end{figure*}
\subsection{Experimental setup \label{sect:expSetup1}}
%\dz{
To identify bound violation patterns and groups of test functions with similar characteristics, we used the following setup:

%\begin{itemize}
 \noindent$\bullet\quad${\it Test functions:} all 24 functions ($n=20$) from instances $1$-$5$ and $101$-$110$ of the SBOX-COST and BBOB test suites;
 
 \noindent$\bullet\quad${\it Methods:} \texttt{DE/rand/1/bin-none} ($N=50$, $F=0.5$, $CR=0.5$); \texttt{L-SHADE-none} ($N_{init}=18n$, $p\in [2/N_{crt},0.2]$ - {\it p}best is selected between the first $p$\% top-ranked individuals, archive size of $6$ \cite{tanabe2014improving});

 \noindent$\bullet\quad${\it Stopping condition:} $n_{fe}=10000n$.
%\end{itemize}

The information collected during the experiments\footnote{The experimental data are available at https://doi.org/10.5281/zenodo.7836831} is the ratio of infeasible components (estimation of bound violation probability, obtained as average on five independent runs). Since the purpose of the experiment is to analyse the characteristics of the test functions and not of the BCHMs, all infeasible elements have been discarded, i.e. similar to the death penalty method and the \texttt{dismiss} strategy in \cite{ISBDE}. In the case of SBOX-COST this is implicitly ensured by the DE selection, as fitness value of infeasible elements is infinite.%}
 
\subsection{Results}
\begin{figure*}[h]
    \centering
    \includegraphics[width=\linewidth,trim=55mm 50mm 25mm 40mm,clip]{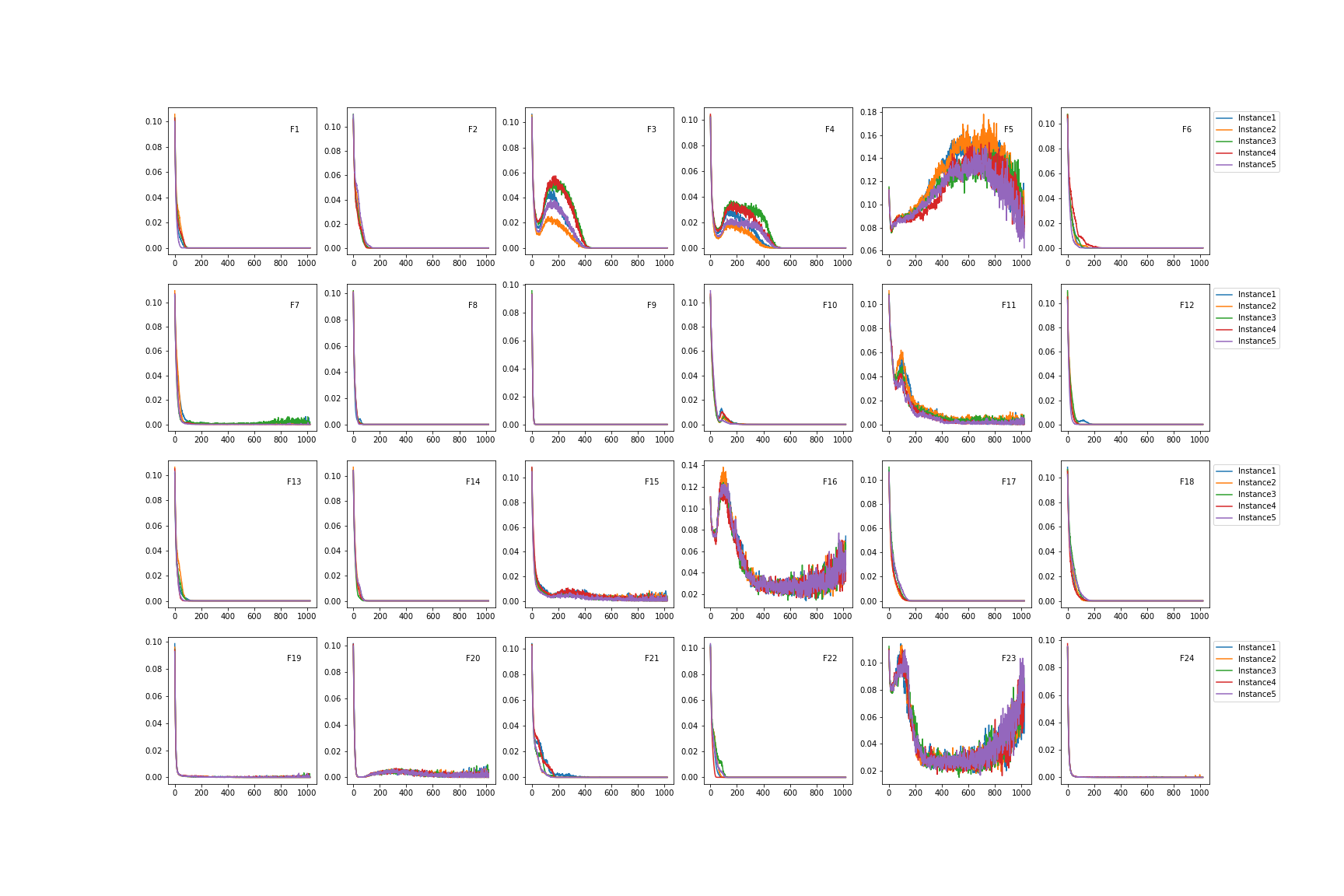}
    \caption{Empirical estimation of bound violation probabilities averaged over 30 runs of L-SHADE (with infeasible individuals discarded) in the case of SBOX-COST functions (instances 1-5).\label{fig:BoundViolProb}}
    \Description{Bound violation probabilities }
\end{figure*}

%\dz{
Figure~\ref{fig:dendrogram} illustrates how the functions of the first $5$ instances of the SBOX-COST and BBOB suites can be grouped with respect to the evolution of the probability of bound violation. The grouping is based on the application of agglomerative clustering on the cosine similarities between the vectors obtained by concatenating the bound violation probability values collected during the number of generations corresponding to the allocated $n_{fe}$ budget.  Functions that are typically clustered together, based on the L-SHADE behaviour (see Fig. \ref{fig:BoundViolProb}), are: 
(i) $\{F5, F16, F23\}$ - the ratio of infeasible components has a nonmonotonous evolution, being around $0.1$ after $1000$ generations;  this would mean that around $10$\% of the trial components are transformed by a BCHM; (ii) $\{F19, F20, F24\}$ - for these functions, the bound violation probability decreases quickly to values smaller than $10^{-2}$, then, during the remaining generations, an infeasible individual is occasionally generated; (iii) $\{F3, F4\}$ - after a quick decrease, the bound violation probability increases, reaches a maximum in the range $0.03-0.06$, then decreases to $0$ by $500$-th generation;  (iv) $\{F12, F13, F14\}$,$\{F17, F18\}$ - the bound violation probability is non-zero in the first $100$-$150$ generations and after that no infeasible individuals are generated.
%\end{itemize}
%}
%\dz{

\begin{figure*}[h]
  \centering
  \includegraphics[width=0.9\linewidth,trim=35mm 67mm 0mm 70mm,clip]{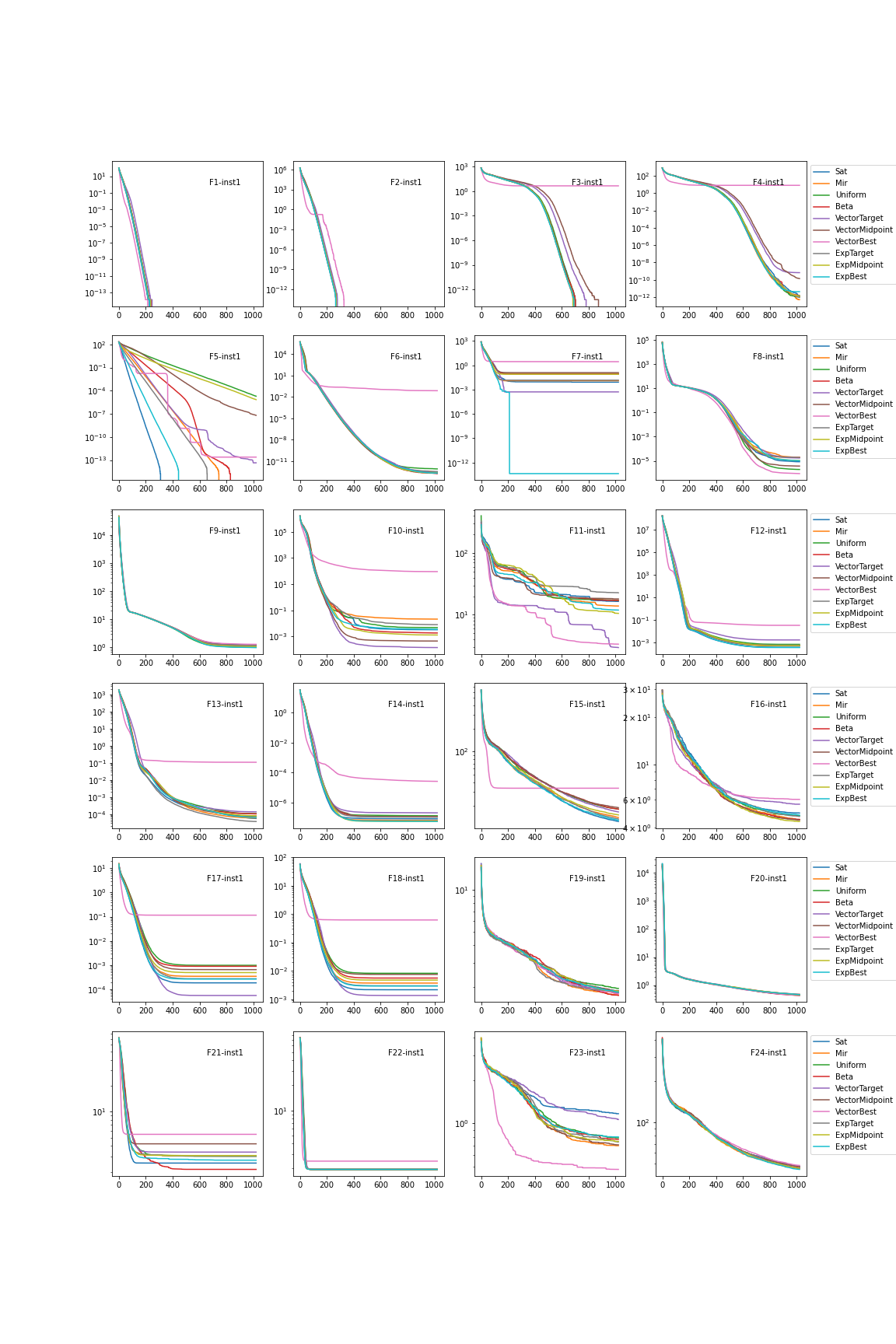}
  \caption{L-SHADE error values (log scale,  averaged over 30 runs,n$_{fe}$=5000n). SBOX-COST functions F1-F24, instance 1.\label{fig:ErrorInst1}}
  \Description{Error values}
\end{figure*}

The grouping patterns identified for the bound violation probabilities can also be observed for the performance values (see the error values for all functions and all analysed BCHMs in Fig.~\ref{fig:ErrorInst1}).%}

%\dz{\hl{@Anna}: are somewhat these groups related to the known properties of the functions?}

\section{Convergence behaviour patterns}
%\dz{
A desirable behaviour of DE is when the population convergence is synchronised with the convergence to the optimum, e.g. the population concentrates on the optimum. This means that when the population diversity becomes small enough, the best element in the population is close enough to the optimum. The case when the population lost all diversity before reaching the optimum corresponds to premature convergence, while the case when the solution is not localised even if the population is still diverse is caused by the inability of the search process to progress toward the solution (one of these behaviours corresponds to stagnation).%}

\subsection{Experimental setup}
To identify the type of behaviour induced by BCHMs when combined with L-SHADE (with the configuration specified in Section \ref{sect:expSetup1}) on SBOX-COST functions, we collected results corresponding to the following cases (all reported results are averages over $5$ independent runs): 

\noindent$\bullet\quad$ {\it Good behaviour (GB).} The difference between the best element value in the population and the optimal value is smaller than $10^{-6}$ and the population variance calculated per component is smaller than $10^{-8}$. This means that the population converged and the solution was localised.

\noindent$\bullet\quad$ {\it Solution found (SF).} The error is smaller than $10^{-6}$, but the population is not convergent (variance greater than $10^{-8}$).

\noindent$\bullet\quad$ {\it Premature convergence (PC).} The variance is smaller than $10^{-8}$ but the error is larger than $10^{-6}$.

\noindent$\bullet\quad$ {\it Bad behaviour (BB).} Both the variance of the population and the error are greater than the thresholds.

\subsection{Results}
\begin{figure*}[h]
  \centering
  \includegraphics[height=0.555\linewidth,trim=30mm 5mm 37mm 9mm,clip]{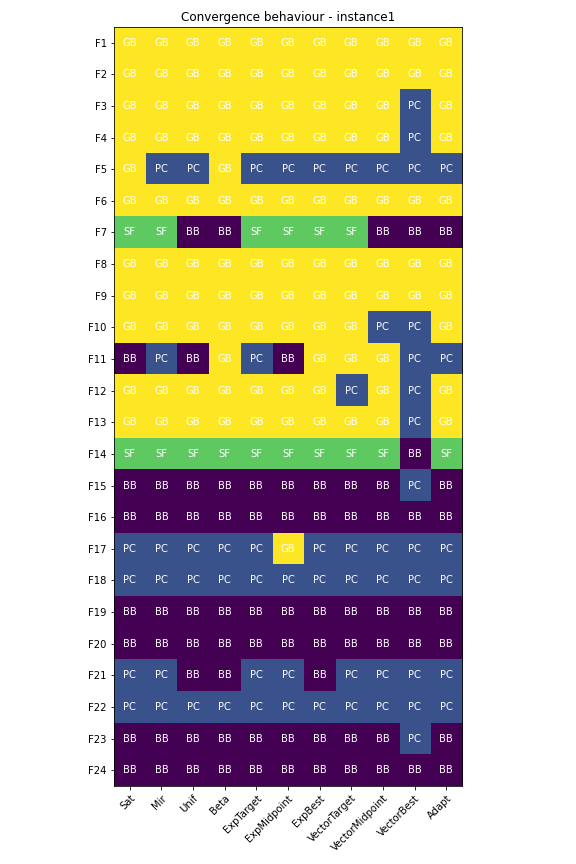}
  \includegraphics[height=0.555\linewidth,trim=40mm 5mm 37mm 9mm,clip] {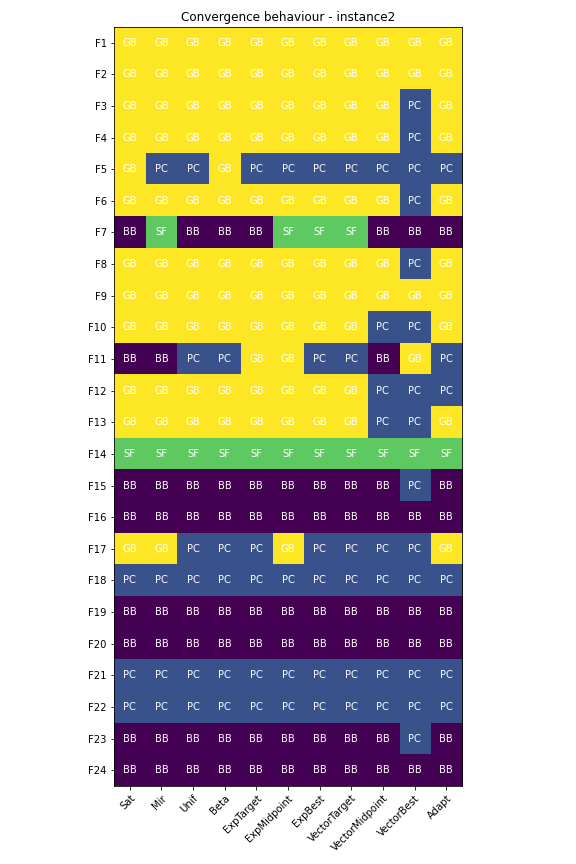}
  \includegraphics[height=0.555\linewidth,trim=40mm 5mm 37mm 9mm,clip] {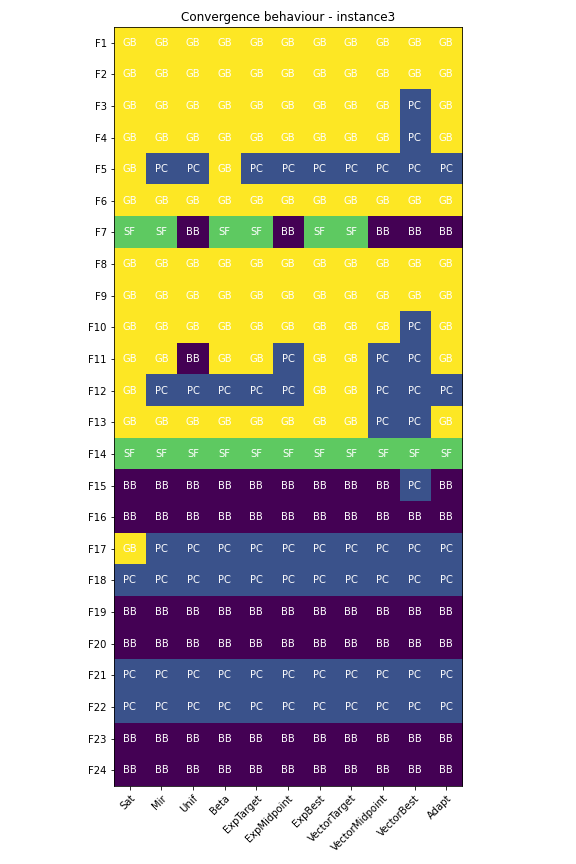}
  \includegraphics[height=0.555\linewidth,trim=40mm 5mm 37mm 9mm,clip] {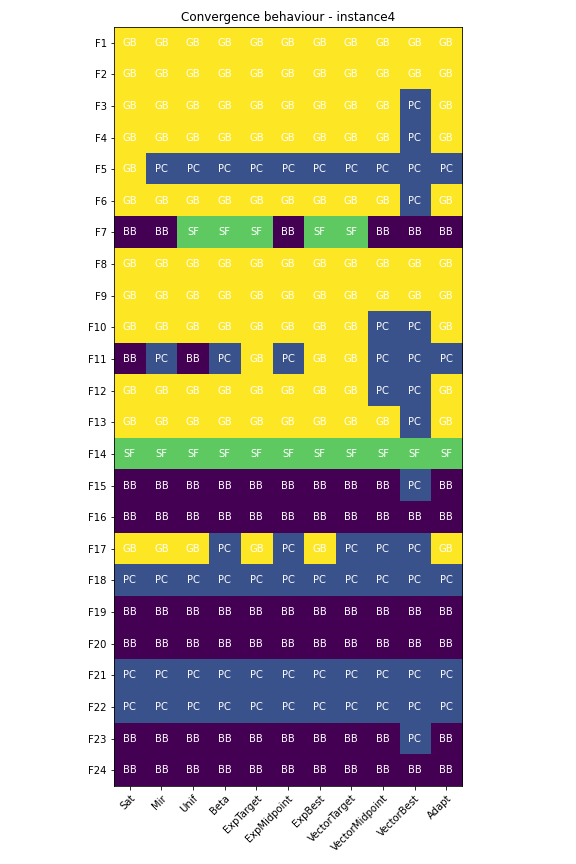}
  \caption{Comparison of BCHMs with respect to the convergence behaviour of L-SHADE on instances 1-4 (left to right): GB (in yellow) means solution localized and convergent population; SF (in green) means solution localized and non-convergent population; PC (in blue) means convergent population but solution not localized, i.e. premature convergence; BB (in violet) means solution not localized and still diverse population. \label{fig:heatmapBCHM}}
  \Description{Heatmap illustrating the behaviour of all BCHMS for all SBOX-COST functions - instances 1, 2, 3, 4.}
\end{figure*}

%\dz{
Figure \ref{fig:heatmapBCHM} illustrates the behavioural differences between the BCHMs analysed together with the test functions. One remark is that the vector-wise BCHMs using the population mean or that using a {\it p}best element are more prone to premature convergence. On the other hand, for the first 14 functions (F1-F14), at least one BCHM leads to the solution, while for the last ten, except for F17, none of the cases is successful.%}  

\section{Towards adaptive BCHM}

\begin{figure*}[h]
    \centering
\includegraphics[width=0.33\linewidth,trim=9mm 0mm 14mm 0mm,clip]{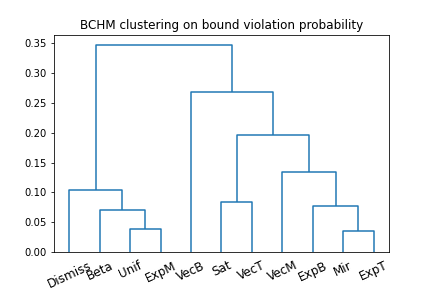}
    \includegraphics[width=0.33\linewidth,trim=9mm 0mm 14mm 0mm,clip]{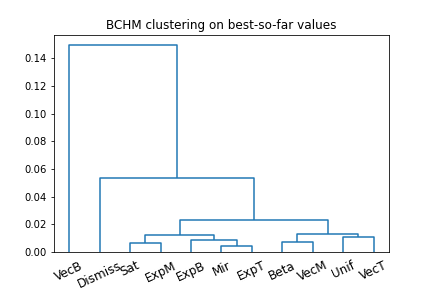}
    \includegraphics[width=0.33\linewidth,trim=9mm 0mm 14mm 0mm,clip]{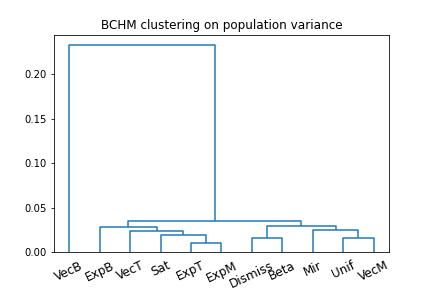}
    \caption{Dendrograms corresponding to agglomerative clustering with complete linkage on cosine similarities between the vectors corresponding to the evolution of (i) estimated bound violation probability (ratio of infeasible components) (ii) best-so-far values; (iii) population variance. Method: L-SHADE; test suite: SBOX-COST (first 5 instances). 
    \label{fig:dendrogram2}}
    \Description{Dendrograms illustrating the clustering of BCHM}
\end{figure*}

%\dz{
The variability in the behaviour of BCHMs for different fitness landscapes makes their choice in the general setting rather difficult. As in the case of other algorithmic components or control parameters, the idea of adaptive selection seems natural. A first approach in this direction was proposed in \cite{Juarez-Castillo2019} where for each infeasible individual, a method is selected from a pool of four BCHMs. The selection is stochastic, being based on a probability distribution adjusted every $25$ generation by using the success ratio of each of the BCHMs in the pool. 
The design of an adaptive BCHM requires: (i) a diverse enough pool of methods; (ii) an adaptive selection strategy.  Figure \ref{fig:dendrogram2} illustrates the cosine similarities between different BCHMs computed using various characteristics (ratio of infeasible components, the evolution of the best-so-far values and the population variance). It can be seen that the results are in line with the particularities of the BCHMs. For example, with respect to the probability of bound violation, \texttt{Sat} is in the same group as vector-wise methods (all of which lead to corrected elements on the boundary). On the other hand \texttt{Beta} correction is in the same group as \texttt{Dismiss} both with respect to the bound violation probability and with respect to population variance (since both aim to preserve the current distribution of the population). As a first step toward the design of an adaptive BCHM, we selected from the groups generated using the best-so-far values (middle dendrogram in Fig.~\ref{fig:dendrogram2}) the following methods: \texttt{VectorBest} (because of its atypical behaviour and the best performance for $F8$, $F22$, and $F23$), \texttt{ExpBest} (best performance for $F4$, $F7$, $F9$, $F10$, $F14$, and $F24$), \texttt{Sat} (best performance for $F5$, $F12$, $F15$), \texttt{VectorTarget} (best performance for $F11$, $F17$, $F18$), \texttt{Beta} (best performance for $F19$, $F20$, $F21$). The performance is assessed by ranking, for each test function, all BCHMs using Glicko2 rating \cite{Glicko2014} from the \texttt{IOHAnalyzer} platform \cite{IOHanalyzer} with $50$ games. The adaptation of the probability distribution used for selection is based on that proposed in \cite{Juarez-Castillo2019}. Preliminary results revealed an average performance of the adaptive variant (see Figure \ref{fig:heatmapBCHM}, last column in each subfigure), suggesting that more analysis is necessary to design an effective adaptive variant.%}

\section{Conclusions}

The conducted analysis revealed the following remarks: (i) for most of the SBOX-COST functions, infeasible individuals are ge\-ne\-rated only in the first generations, leading to a negligible impact of BCHMs (e.g. $F9$); (ii) there are functions (e.g. $F16$, $F23$) with non-zero bound violation probability even after $1000$ generations; (iii) the guiding role of the BCHM, ensured by using a reference vector, is important (\texttt{ExpBest}, \texttt{ExpTarget}, \texttt{VectBest}, \texttt{VectTarget} led to best results in half of the SBOX-COST functions); (iv) preservation of the DE population or search characteristics (specific to \texttt{Beta} and \texttt{VectTarget}) proved to be useful for some functions ($F21$; $F11$, $F17$, $F18$). Finally, when considering BCHMs as algorithmic components of search metaheuristics, we believe that identifying guidelines for the design of adaptive correction methods is a path to follow.

\balance

\bibliographystyle{ACM-Reference-Format}
\bibliography{main}


\begin{thebibliography}{29}

%%% the Web address.

\ifx \showCODEN    \undefined \def \showCODEN     #1{\unskip}     \fi
\ifx \showDOI      \undefined \def \showDOI       #1{#1}\fi
\ifx \showISBNx    \undefined \def \showISBNx     #1{\unskip}     \fi
\ifx \showISBNxiii \undefined \def \showISBNxiii  #1{\unskip}     \fi
\ifx \showISSN     \undefined \def \showISSN      #1{\unskip}     \fi
\ifx \showLCCN     \undefined \def \showLCCN      #1{\unskip}     \fi
\ifx \shownote     \undefined \def \shownote      #1{#1}          \fi
\ifx \showarticletitle \undefined \def \showarticletitle #1{#1}   \fi
\ifx \showURL      \undefined \def \showURL       {\relax}        \fi
% invisible to TeX
\providecommand\bibfield[2]{#2}
\providecommand\bibinfo[2]{#2}
\providecommand\natexlab[1]{#1}
\providecommand\showeprint[2][]{arXiv:#2}

\bibitem[\protect\citeauthoryear{Ali and Fatti}{Ali and Fatti}{2006}]%
        {Ali2006}
\bibfield{author}{\bibinfo{person}{M.~M. Ali} {and} \bibinfo{person}{L.~P.
  Fatti}.} \bibinfo{year}{2006}\natexlab{}.
\newblock \showarticletitle{A Differential Free Point Generation Scheme in the
  Differential Evolution Algorithm}.
\newblock \bibinfo{journal}{\emph{J. Glob. Optim.}} \bibinfo{volume}{35},
  \bibinfo{number}{4} (\bibinfo{year}{2006}), \bibinfo{pages}{551--572}.
\newblock
\urldef\tempurl%
\url{https://doi.org/10.1007/s10898-005-3767-y}
\showDOI{\tempurl}


\bibitem[\protect\citeauthoryear{Arabas, Szczepankiewicz, and Wroniak}{Arabas
  et~al\mbox{.}}{2010}]%
        {Arabas2010}
\bibfield{author}{\bibinfo{person}{Jaroslaw Arabas}, \bibinfo{person}{Adam
  Szczepankiewicz}, {and} \bibinfo{person}{Tomasz Wroniak}.}
  \bibinfo{year}{2010}\natexlab{}.
\newblock \showarticletitle{Experimental Comparison of Methods to Handle
  Boundary Constraints in Differential Evolution}.
\newblock   \bibinfo{volume}{6239} (\bibinfo{year}{2010}),
  \bibinfo{pages}{411--420}.
\newblock
\urldef\tempurl%
\url{https://doi.org/doi.org/10.1007/978-3-642-15871-1\_42}
\showDOI{\tempurl}


\bibitem[\protect\citeauthoryear{Biedrzycki}{Biedrzycki}{2020}]%
        {Biedrzycki2020}
\bibfield{author}{\bibinfo{person}{Rafał Biedrzycki}.}
  \bibinfo{year}{2020}\natexlab{}.
\newblock \showarticletitle{Handling bound constraints in CMA-ES: An
  experimental study}.
\newblock \bibinfo{journal}{\emph{Swarm and Evolutionary Computation}}
  \bibinfo{volume}{52} (\bibinfo{year}{2020}), \bibinfo{pages}{100627}.
\newblock
\showISSN{2210-6502}
\urldef\tempurl%
\url{https://doi.org/10.1016/j.swevo.2019.100627}
\showDOI{\tempurl}


\bibitem[\protect\citeauthoryear{Biedrzycki, Arabas, and
  Jagodzinski}{Biedrzycki et~al\mbox{.}}{2019}]%
        {Biedrzycki2019}
\bibfield{author}{\bibinfo{person}{Rafal Biedrzycki}, \bibinfo{person}{Jaroslaw
  Arabas}, {and} \bibinfo{person}{Dariusz Jagodzinski}.}
  \bibinfo{year}{2019}\natexlab{}.
\newblock \showarticletitle{Bound constraints handling in Differential
  Evolution: An experimental study}.
\newblock \bibinfo{journal}{\emph{Swarm Evol. Comput.}}  \bibinfo{volume}{50}
  (\bibinfo{year}{2019}).
\newblock
\urldef\tempurl%
\url{https://doi.org/10.1016/j.swevo.2018.10.004}
\showDOI{\tempurl}


\bibitem[\protect\citeauthoryear{Boks, Kononova, and Wang}{Boks
  et~al\mbox{.}}{2021}]%
        {Boks2021}
\bibfield{author}{\bibinfo{person}{Rick Boks}, \bibinfo{person}{Anna~V.
  Kononova}, {and} \bibinfo{person}{Hao Wang}.}
  \bibinfo{year}{2021}\natexlab{}.
\newblock \showarticletitle{Quantifying the impact of boundary constraint
  handling methods on differential evolution}. In
  \bibinfo{booktitle}{\emph{{GECCO} '21: Genetic and Evolutionary Computation
  Conference, Companion Volume, Lille, France, July 10-14, 2021}},
  \bibfield{editor}{\bibinfo{person}{Krzysztof Krawiec}} (Ed.).
  \bibinfo{publisher}{{ACM}}, \bibinfo{pages}{1199--1207}.
\newblock
\urldef\tempurl%
\url{https://doi.org/10.1145/3449726.3463214}
\showDOI{\tempurl}


\bibitem[\protect\citeauthoryear{Brest and Mau\v{c}ec}{Brest and
  Mau\v{c}ec}{2008}]%
        {bib:Brest2008b}
\bibfield{author}{\bibinfo{person}{Janez Brest} {and}
  \bibinfo{person}{Mirjam~Sepesy Mau\v{c}ec}.} \bibinfo{year}{2008}\natexlab{}.
\newblock \showarticletitle{Population size reduction for the differential
  evolution algorithm}.
\newblock \bibinfo{journal}{\emph{Applied Intelligence}} \bibinfo{volume}{29},
  \bibinfo{number}{3} (\bibinfo{year}{2008}), \bibinfo{pages}{228--247}.
\newblock


\bibitem[\protect\citeauthoryear{Caraffini, Kononova, and Corne}{Caraffini
  et~al\mbox{.}}{2019}]%
        {ISBDE}
\bibfield{author}{\bibinfo{person}{Fabio Caraffini}, \bibinfo{person}{Anna~V.
  Kononova}, {and} \bibinfo{person}{David Corne}.}
  \bibinfo{year}{2019}\natexlab{}.
\newblock \showarticletitle{Infeasibility and structural bias in differential
  evolution}.
\newblock \bibinfo{journal}{\emph{Information Sciences}}  \bibinfo{volume}{496}
  (\bibinfo{year}{2019}), \bibinfo{pages}{161--179}.
\newblock
\showISSN{0020-0255}
\urldef\tempurl%
\url{https://doi.org/10.1016/j.ins.2019.05.019}
\showDOI{\tempurl}


\bibitem[\protect\citeauthoryear{Das, Mullick, and Suganthan}{Das
  et~al\mbox{.}}{2016}]%
        {DAS20161}
\bibfield{author}{\bibinfo{person}{S. Das}, \bibinfo{person}{Sankha~Subhra
  Mullick}, {and} \bibinfo{person}{P.N. Suganthan}.}
  \bibinfo{year}{2016}\natexlab{}.
\newblock \showarticletitle{Recent advances in differential evolution – An
  updated survey}.
\newblock \bibinfo{journal}{\emph{Swarm and Evolutionary Computation}}
  \bibinfo{volume}{27} (\bibinfo{year}{2016}), \bibinfo{pages}{1 -- 30}.
\newblock
\urldef\tempurl%
\url{https://doi.org/10.1016/j.swevo.2016.01.004}
\showDOI{\tempurl}


\bibitem[\protect\citeauthoryear{de{-}la{-}Cruz{-}Mart{\'{\i}}nez and
  Mezura{-}Montes}{de{-}la{-}Cruz{-}Mart{\'{\i}}nez and
  Mezura{-}Montes}{2020}]%
        {Martinez2020}
\bibfield{author}{\bibinfo{person}{Sebasti{\'{a}}n{-}Jos{\'{e}}
  de{-}la{-}Cruz{-}Mart{\'{\i}}nez} {and} \bibinfo{person}{Efr{\'{e}}n
  Mezura{-}Montes}.} \bibinfo{year}{2020}\natexlab{}.
\newblock \showarticletitle{Boundary Constraint-Handling Methods in
  Differential Evolution for Mechanical Design Optimization}. In
  \bibinfo{booktitle}{\emph{{IEEE} Congress on Evolutionary Computation
  ({CEC})}}. \bibinfo{publisher}{{IEEE}}, \bibinfo{pages}{1--8}.
\newblock
\urldef\tempurl%
\url{https://doi.org/10.1109/CEC48606.2020.9185495}
\showDOI{\tempurl}


\bibitem[\protect\citeauthoryear{Doerr, Wang, Ye, van Rijn, and B{\"a}ck}{Doerr
  et~al\mbox{.}}{2018}]%
        {IOHprofiler}
\bibfield{author}{\bibinfo{person}{Carola Doerr}, \bibinfo{person}{Hao Wang},
  \bibinfo{person}{Furong Ye}, \bibinfo{person}{Sander van Rijn}, {and}
  \bibinfo{person}{Thomas B{\"a}ck}.} \bibinfo{year}{2018}\natexlab{}.
\newblock \showarticletitle{IOHprofiler: A Benchmarking and Profiling Tool for
  Iterative Optimization Heuristics}.
\newblock \bibinfo{journal}{\emph{arXiv e-prints:1810.05281}}
  (\bibinfo{date}{oct} \bibinfo{year}{2018}).
\newblock
\showeprint[arxiv]{1810.05281}
\urldef\tempurl%
\url{https://arxiv.org/abs/1810.05281}
\showURL{%
\tempurl}


\bibitem[\protect\citeauthoryear{Hansen, Auger, Ros, Mersmann, Tušar, and
  Brockhoff}{Hansen et~al\mbox{.}}{2021}]%
        {BBOB2021}
\bibfield{author}{\bibinfo{person}{Nikolaus Hansen}, \bibinfo{person}{Anne
  Auger}, \bibinfo{person}{Raymond Ros}, \bibinfo{person}{Olaf Mersmann},
  \bibinfo{person}{Tea Tušar}, {and} \bibinfo{person}{Dimo Brockhoff}.}
  \bibinfo{year}{2021}\natexlab{}.
\newblock \showarticletitle{COCO: a platform for comparing continuous
  optimizers in a black-box setting}.
\newblock \bibinfo{journal}{\emph{Optimization Methods and Software}}
  \bibinfo{volume}{36}, \bibinfo{number}{1} (\bibinfo{year}{2021}),
  \bibinfo{pages}{114--144}.
\newblock
\urldef\tempurl%
\url{https://doi.org/10.1080/10556788.2020.1808977}
\showDOI{\tempurl}
\showeprint{https://doi.org/10.1080/10556788.2020.1808977}


\bibitem[\protect\citeauthoryear{Helwig, Branke, and Mostaghim}{Helwig
  et~al\mbox{.}}{2013}]%
        {Helwig2013}
\bibfield{author}{\bibinfo{person}{Sabine Helwig},
  \bibinfo{person}{J{\"{u}}rgen Branke}, {and} \bibinfo{person}{Sanaz
  Mostaghim}.} \bibinfo{year}{2013}\natexlab{}.
\newblock \showarticletitle{Experimental Analysis of Bound Handling Techniques
  in Particle Swarm Optimization}.
\newblock \bibinfo{journal}{\emph{{IEEE} Transactions on Evolutionary
  Computation}} \bibinfo{volume}{17}, \bibinfo{number}{2}
  (\bibinfo{year}{2013}), \bibinfo{pages}{259--271}.
\newblock
\urldef\tempurl%
\url{https://doi.org/10.1109/TEVC.2012.2189404}
\showDOI{\tempurl}


\bibitem[\protect\citeauthoryear{Helwig and Wanka}{Helwig and Wanka}{2008}]%
        {Helwig2008}
\bibfield{author}{\bibinfo{person}{Sabine Helwig} {and} \bibinfo{person}{Rolf
  Wanka}.} \bibinfo{year}{2008}\natexlab{}.
\newblock \showarticletitle{Theoretical Analysis of Initial Particle Swarm
  Behavior}. In \bibinfo{booktitle}{\emph{Parallel Problem Solving from Nature
  -- PPSN X}}, \bibfield{editor}{\bibinfo{person}{G{\"u}nter Rudolph},
  \bibinfo{person}{Thomas Jansen}, \bibinfo{person}{Nicola Beume},
  \bibinfo{person}{Simon Lucas}, {and} \bibinfo{person}{Carlo Poloni}} (Eds.).
  \bibinfo{publisher}{Springer Berlin Heidelberg}, \bibinfo{address}{Berlin,
  Heidelberg}, \bibinfo{pages}{889--898}.
\newblock


\bibitem[\protect\citeauthoryear{Ju{\'{a}}rez{-}Castillo, Acosta{-}Mesa, and
  Mezura{-}Montes}{Ju{\'{a}}rez{-}Castillo et~al\mbox{.}}{2017}]%
        {Castillo2017}
\bibfield{author}{\bibinfo{person}{Efr{\'{e}}n Ju{\'{a}}rez{-}Castillo},
  \bibinfo{person}{H{\'{e}}ctor{-}Gabriel Acosta{-}Mesa}, {and}
  \bibinfo{person}{Efr{\'{e}}n Mezura{-}Montes}.}
  \bibinfo{year}{2017}\natexlab{}.
\newblock \showarticletitle{Empirical study of bound constraint-handling
  methods in Particle Swarm Optimization for constrained search spaces}. In
  \bibinfo{booktitle}{\emph{2017 {IEEE} Congress on Evolutionary Computation}}.
  \bibinfo{publisher}{{IEEE}}, \bibinfo{pages}{604--611}.
\newblock
\urldef\tempurl%
\url{https://doi.org/10.1109/CEC.2017.7969366}
\showDOI{\tempurl}


\bibitem[\protect\citeauthoryear{Ju{\'{a}}rez{-}Castillo, Acosta{-}Mesa, and
  Mezura{-}Montes}{Ju{\'{a}}rez{-}Castillo et~al\mbox{.}}{2019}]%
        {Juarez-Castillo2019}
\bibfield{author}{\bibinfo{person}{Efr{\'{e}}n Ju{\'{a}}rez{-}Castillo},
  \bibinfo{person}{H{\'{e}}ctor{-}Gabriel Acosta{-}Mesa}, {and}
  \bibinfo{person}{Efr{\'{e}}n Mezura{-}Montes}.}
  \bibinfo{year}{2019}\natexlab{}.
\newblock \showarticletitle{Adaptive boundary constraint-handling scheme for
  constrained optimization}.
\newblock \bibinfo{journal}{\emph{Soft Comput.}} \bibinfo{volume}{23},
  \bibinfo{number}{17} (\bibinfo{year}{2019}), \bibinfo{pages}{8247--8280}.
\newblock
\urldef\tempurl%
\url{https://doi.org/10.1007/s00500-018-3459-4}
\showDOI{\tempurl}


\bibitem[\protect\citeauthoryear{Kadavy, Viktorin, Kazikova, Pluhacek, and
  Senkerik}{Kadavy et~al\mbox{.}}{2022}]%
        {Kadavy2022}
\bibfield{author}{\bibinfo{person}{Tomas Kadavy}, \bibinfo{person}{Adam
  Viktorin}, \bibinfo{person}{Anezka Kazikova}, \bibinfo{person}{Michal
  Pluhacek}, {and} \bibinfo{person}{Roman Senkerik}.}
  \bibinfo{year}{2022}\natexlab{}.
\newblock \showarticletitle{Impact of Boundary Control Methods on
  Bound-Constrained Optimization Benchmarking}.
\newblock \bibinfo{journal}{\emph{IEEE Transactions on Evolutionary
  Computation}} \bibinfo{volume}{26}, \bibinfo{number}{6}
  (\bibinfo{year}{2022}), \bibinfo{pages}{1271--1280}.
\newblock
\urldef\tempurl%
\url{https://doi.org/10.1109/TEVC.2022.3204412}
\showDOI{\tempurl}


\bibitem[\protect\citeauthoryear{Kononova, Caraffini, and Bäck}{Kononova
  et~al\mbox{.}}{2021}]%
        {KONONOVA2021}
\bibfield{author}{\bibinfo{person}{Anna~V. Kononova}, \bibinfo{person}{Fabio
  Caraffini}, {and} \bibinfo{person}{Thomas Bäck}.}
  \bibinfo{year}{2021}\natexlab{}.
\newblock \showarticletitle{Differential evolution outside the box}.
\newblock \bibinfo{journal}{\emph{Information Sciences}}  \bibinfo{volume}{581}
  (\bibinfo{year}{2021}), \bibinfo{pages}{587--604}.
\newblock
\showISSN{0020-0255}
\urldef\tempurl%
\url{https://doi.org/10.1016/j.ins.2021.09.058}
\showDOI{\tempurl}


\bibitem[\protect\citeauthoryear{Kononova, Vermetten, Caraffini, Mitran, and
  Zaharie}{Kononova et~al\mbox{.}}{2022}]%
        {Kononova2022importance}
\bibfield{author}{\bibinfo{person}{Anna~V. Kononova},
  \bibinfo{person}{Diederick Vermetten}, \bibinfo{person}{Fabio Caraffini},
  \bibinfo{person}{Madalina-A. Mitran}, {and} \bibinfo{person}{Daniela
  Zaharie}.} \bibinfo{year}{2022}\natexlab{}.
\newblock \bibinfo{title}{The importance of being constrained: dealing with
  infeasible solutions in Differential Evolution and beyond}.
\newblock
\newblock
\showeprint[arxiv]{cs.NE/2203.03512}


\bibitem[\protect\citeauthoryear{Kreischer, Magalhaes, Barbosa, and
  Krempser}{Kreischer et~al\mbox{.}}{2017}]%
        {kreischer2017}
\bibfield{author}{\bibinfo{person}{Vinicius Kreischer},
  \bibinfo{person}{Thiago~Tavares Magalhaes}, \bibinfo{person}{HJ Barbosa},
  {and} \bibinfo{person}{Eduardo Krempser}.} \bibinfo{year}{2017}\natexlab{}.
\newblock \showarticletitle{Evaluation of bound constraints handling methods in
  differential evolution using the cec2017 benchmark}. In
  \bibinfo{booktitle}{\emph{XIII Brazilian Congress on Computational
  Intelligence}}.
\newblock


\bibitem[\protect\citeauthoryear{Oldewage, Engelbrecht, and Cleghorn}{Oldewage
  et~al\mbox{.}}{2018}]%
        {Oldewage2018}
\bibfield{author}{\bibinfo{person}{Elre~T. Oldewage},
  \bibinfo{person}{Andries~P. Engelbrecht}, {and}
  \bibinfo{person}{Christopher~Wesley Cleghorn}.}
  \bibinfo{year}{2018}\natexlab{}.
\newblock \showarticletitle{Boundary Constraint Handling Techniques for
  Particle Swarm Optimization in High Dimensional Problem Spaces}. In
  \bibinfo{booktitle}{\emph{Swarm Intelligence - 11th International
  Conference}} \emph{(\bibinfo{series}{Lecture Notes in Computer Science})},
  Vol.~\bibinfo{volume}{11172}. \bibinfo{pages}{333--341}.
\newblock
\urldef\tempurl%
\url{https://doi.org/10.1007/978-3-030-00533-7\_27}
\showDOI{\tempurl}


\bibitem[\protect\citeauthoryear{Padhye, Mittal, and Deb}{Padhye
  et~al\mbox{.}}{2015}]%
        {Padhye2015}
\bibfield{author}{\bibinfo{person}{Nikhil Padhye}, \bibinfo{person}{Pulkit
  Mittal}, {and} \bibinfo{person}{Kalyanmoy Deb}.}
  \bibinfo{year}{2015}\natexlab{}.
\newblock \showarticletitle{Feasibility preserving constraint-handling
  strategies for real parameter evolutionary optimization}.
\newblock \bibinfo{journal}{\emph{Computational Optimization and Applications}}
   \bibinfo{volume}{62} (\bibinfo{year}{2015}), \bibinfo{pages}{851--890}.
\newblock
\urldef\tempurl%
\url{https://doi.org/doi.org/10.1007/s10589-015-9752-6}
\showDOI{\tempurl}


\bibitem[\protect\citeauthoryear{Price, Storn, and Lampinen}{Price
  et~al\mbox{.}}{2005}]%
        {bib:DEbook}
\bibfield{author}{\bibinfo{person}{Kenneth~V. Price}, \bibinfo{person}{Rainer
  Storn}, {and} \bibinfo{person}{Jouni Lampinen}.}
  \bibinfo{year}{2005}\natexlab{}.
\newblock \bibinfo{booktitle}{\emph{Differential Evolution: A Practical
  Approach to Global Optimization}}.
\newblock \bibinfo{publisher}{Springer}, \bibinfo{address}{Berlin, Heidelberg}.
\newblock
\urldef\tempurl%
\url{https://doi.org/10.1007/3-540-31306-0}
\showDOI{\tempurl}


\bibitem[\protect\citeauthoryear{Suganthan}{Suganthan}{[n. d.]}]%
        {CEC}
\bibfield{author}{\bibinfo{person}{Ponnuthurai~Nagaratnam Suganthan}.}
  \bibinfo{year}{[n. d.]}\natexlab{}.
\newblock \bibinfo{title}{Benchmarks for Evaluation of Evolutionary
  Algorithms}.
\newblock
  \bibinfo{howpublished}{\url{https://www3.ntu.edu.sg/home/epnsugan/index_files/cec-benchmarking.htm}}.
\newblock
\newblock
\shownote{Accessed: 2023-04-10.}


\bibitem[\protect\citeauthoryear{Tanabe and Fukunaga}{Tanabe and
  Fukunaga}{2013}]%
        {bib:tanabe2013}
\bibfield{author}{\bibinfo{person}{Ryoji Tanabe} {and} \bibinfo{person}{Alex
  Fukunaga}.} \bibinfo{year}{2013}\natexlab{}.
\newblock \showarticletitle{Success-history based parameter adaptation for
  Differential Evolution}. In \bibinfo{booktitle}{\emph{2013 IEEE Congress on
  Evolutionary Computation}}. \bibinfo{pages}{71--78}.
\newblock
\urldef\tempurl%
\url{https://doi.org/10.1109/CEC.2013.6557555}
\showDOI{\tempurl}


\bibitem[\protect\citeauthoryear{Tanabe and Fukunaga}{Tanabe and
  Fukunaga}{2014}]%
        {tanabe2014improving}
\bibfield{author}{\bibinfo{person}{Ryoji Tanabe} {and} \bibinfo{person}{Alex~S
  Fukunaga}.} \bibinfo{year}{2014}\natexlab{}.
\newblock \showarticletitle{Improving the search performance of SHADE using
  linear population size reduction}. In \bibinfo{booktitle}{\emph{2014 IEEE
  Congress on Evolutionary Computation (CEC)}}. IEEE,
  \bibinfo{publisher}{IEEE}, \bibinfo{pages}{1658--1665}.
\newblock


\bibitem[\protect\citeauthoryear{Vecek, Mernik, and Crepinsek}{Vecek
  et~al\mbox{.}}{2014}]%
        {Glicko2014}
\bibfield{author}{\bibinfo{person}{Niki Vecek}, \bibinfo{person}{Marjan
  Mernik}, {and} \bibinfo{person}{Matej Crepinsek}.}
  \bibinfo{year}{2014}\natexlab{}.
\newblock \showarticletitle{A chess rating system for evolutionary algorithms:
  {A} new method for the comparison and ranking of evolutionary algorithms}.
\newblock \bibinfo{journal}{\emph{Inf. Sci.}}  \bibinfo{volume}{277}
  (\bibinfo{year}{2014}), \bibinfo{pages}{656--679}.
\newblock
\urldef\tempurl%
\url{https://doi.org/10.1016/j.ins.2014.02.154}
\showDOI{\tempurl}


\bibitem[\protect\citeauthoryear{Wang, Vermetten, Ye, Doerr, and B\"{a}ck}{Wang
  et~al\mbox{.}}{2022}]%
        {IOHanalyzer}
\bibfield{author}{\bibinfo{person}{Hao Wang}, \bibinfo{person}{Diederick
  Vermetten}, \bibinfo{person}{Furong Ye}, \bibinfo{person}{Carola Doerr},
  {and} \bibinfo{person}{Thomas B\"{a}ck}.} \bibinfo{year}{2022}\natexlab{}.
\newblock \showarticletitle{IOHanalyzer: Detailed Performance Analyses for
  Iterative Optimization Heuristics}.
\newblock \bibinfo{journal}{\emph{ACM Trans. Evol. Learn. Optim.}}
  \bibinfo{volume}{2}, \bibinfo{number}{1}, Article \bibinfo{articleno}{3}
  (\bibinfo{date}{apr} \bibinfo{year}{2022}), \bibinfo{numpages}{29}~pages.
\newblock
\showISSN{2688-299X}
\urldef\tempurl%
\url{https://doi.org/10.1145/3510426}
\showDOI{\tempurl}


\bibitem[\protect\citeauthoryear{Wessing}{Wessing}{2013}]%
        {Wessing2013}
\bibfield{author}{\bibinfo{person}{Simon Wessing}.}
  \bibinfo{year}{2013}\natexlab{}.
\newblock \bibinfo{booktitle}{\emph{Repair Methods for Box Constraints
  Revisited}}. \bibinfo{series}{Lecture Notes in Computer Science},
  Vol.~\bibinfo{volume}{7835}.
\newblock \bibinfo{publisher}{Springer}, \bibinfo{address}{Berlin, Heidelberg},
  \bibinfo{pages}{469--478}.
\newblock
\urldef\tempurl%
\url{https://doi.org/10.1007/978-3-642-37192-9\_47}
\showURL{%
\tempurl}


\bibitem[\protect\citeauthoryear{Zhang and Sanderson}{Zhang and
  Sanderson}{2009}]%
        {bib:zhang2009}
\bibfield{author}{\bibinfo{person}{Jingqiao Zhang} {and}
  \bibinfo{person}{Arthur~C. Sanderson}.} \bibinfo{year}{2009}\natexlab{}.
\newblock \showarticletitle{{JADE}: Adaptive Differential Evolution With
  Optional External Archive}.
\newblock \bibinfo{journal}{\emph{IEEE Transactions on Evolutionary
  Computation}} \bibinfo{volume}{13}, \bibinfo{number}{5}
  (\bibinfo{year}{2009}), \bibinfo{pages}{945--958}.
\newblock
\urldef\tempurl%
\url{https://doi.org/10.1109/TEVC.2009.2014613}
\showDOI{\tempurl}


\end{thebibliography}

\end{document}